\documentclass[10pt,twocolumn,letterpaper]{article}

\usepackage{iccv}              
\definecolor{iccvblue}{rgb}{0.21,0.49,0.74}
\usepackage[pagebackref,breaklinks,colorlinks,allcolors=iccvblue]{hyperref}

\def\paperID{7812} 
\def\confName{ICCV}
\def\confYear{2025}

\title{MamV2XCalib: V2X-based Target-less Infrastructure Camera Calibration with State Space Model}

\author{
Yaoye Zhu \quad Zhe Wang \quad Yan Wang\thanks{Corresponding author: \texttt{wangyan@air.tsinghua.edu.cn}}\\
Institute for AI Industry Research (AIR), Tsinghua University\\
{\tt\small zhuyaoye22@gmail.com \quad wangzhe@air.tsinghua.edu.cn \quad wangyan@air.tsinghua.edu.cn}
}
\begin{document}
\maketitle
\begin{abstract}
As cooperative systems that leverage roadside cameras to assist autonomous vehicle perception become increasingly widespread, large-scale precise calibration of infrastructure cameras has become a critical issue. Traditional manual calibration methods are often time-consuming, labor-intensive, and may require road closures.  This paper proposes MamV2XCalib, the first V2X-based infrastructure camera calibration method with the assistance of vehicle-side LiDAR. MamV2XCalib only requires autonomous vehicles equipped with LiDAR to drive near the cameras to be calibrated in the infrastructure, without the need for specific reference objects or manual intervention. We also introduce a new targetless LiDAR-camera calibration method, which combines multi-scale features and a 4D correlation volume to estimate the correlation between vehicle-side point clouds and roadside images. We model the temporal information and estimate the rotation angles with Mamba, effectively addressing calibration failures in V2X scenarios caused by defects in the vehicle-side data (such as occlusions) and large differences in viewpoint. We evaluate MamV2XCalib on the V2X-Seq and TUMTraf-V2X real-world datasets, demonstrating the effectiveness and robustness of our V2X-based automatic calibration approach. Compared to previous LiDAR-camera methods designed for calibration on one car, our approach achieves better and more stable calibration performance in V2X scenarios with fewer parameters. The code is available at https://github.com/zhuyaoye/MamV2XCalib.

\end{abstract}
\section{Introduction}
\label{sec:intro}

\begin{figure}[htbp]
    \centering
    \begin{minipage}{0.48\linewidth}
        \centering
        \includegraphics[width=\linewidth]{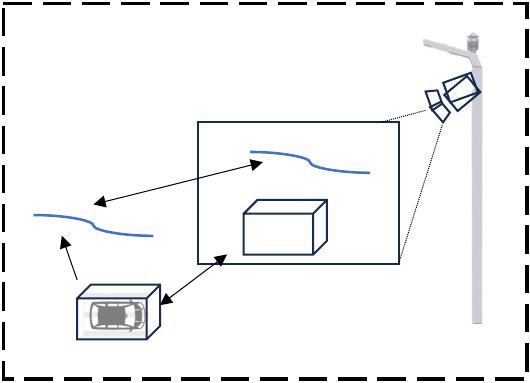}
        \subcaption{Calibration with targets including trajectory or vehicle itself}
        \label{fig:fig1a}
    \end{minipage}
    \begin{minipage}{0.48\linewidth}
        \centering
        \includegraphics[width=\linewidth]{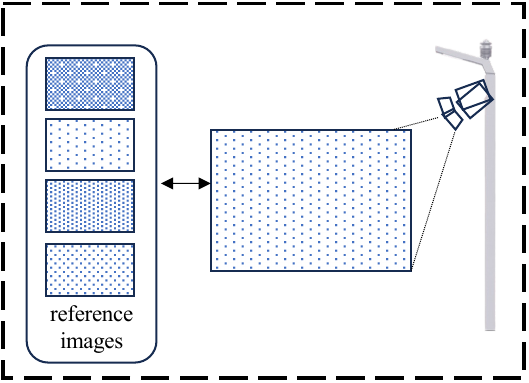}
        \subcaption{Calibration with reference images prepared before calibration}
        \label{fig:fig1b}
    \end{minipage}
    
    \begin{minipage}{0.95\linewidth}
        \centering
        \includegraphics[width=\linewidth]{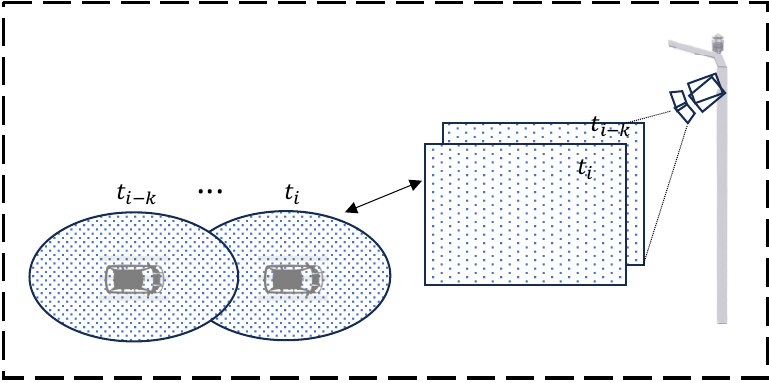}
        \subcaption{Ours: V2X-based target-less camera calibration method}
        \label{fig:fig1c}
    \end{minipage}
    \caption{Comparison of roadside camera calibration methods.}
    \label{fig:figure1}
\end{figure}

In recent years, due to the inherent limitations of on-board perception systems, many researchers have started focusing on utilizing external sensor information to assist autonomous vehicles~\cite{cui2022coopernaut,huang2022rd}. Among these external sources, infrastructure cameras are widely used in V2X systems for their wide deployment, broader field of view, and reduced occlusion~\cite{rinner2008introduction}. However, outdoor deployment exposes roadside cameras to extreme weather and vibrations from large vehicles, causing shifts from their initial positions. If not recalibrated, such shifts can significantly degrade the accuracy of data fusion between road and vehicle sensors. In cases of large deviations, they may even result in a complete loss of monitoring over the intended area~\cite{zhang2024cooperative}. Therefore, efficient, scalable calibration methods are essential for stable intelligent transportation systems in outdoor environments.

Traditional calibration methods~\cite{zhang2000flexible,wang2007research} primarily rely on manual targets, but unlike vehicle sensors that can be calibrated indoors, fixed roadside cameras cannot be relocated. As the demand for camera calibration increases, these manual methods become time-consuming and labor-intensive and may cause disruptions to regular traffic operations. Recent efforts focus on automated calibration~\cite{dubska2014fully,tsaregorodtsev2023automated,zhao2024graph}. Some methods use autonomous vehicle trajectories or vehicle detection targets. However, these approaches heavily depend on the accuracy of trajectory prediction and object detection, while overlooking rich background data. Other methods require the preparation of reference images or use real-world map images, employing virtual cameras aligned with these images as reference objects~\cite{d2024automated}. Preparing a series of reference images for each camera is both inconvenient and lacks real-time adaptability.

Whether traditional or modern methods, determining the position of a sensor in the world coordinate system requires two key conditions: known reference objects and the sensor’s perception of these objects. In road scenarios, autonomous vehicles, being high-quality intelligent agents, possess real-time, robust capabilities to collect environmental data and accurately localize themselves. \textit{Given this, can we invert the typical V2X process, using a vehicle’s perception data to treat the entire environment as a reference, thereby calibrating multiple roadside cameras as the vehicle moves?}

We propose a novel V2X-based calibration algorithm that fuses vehicle-mounted LiDAR point clouds with roadside camera images. A neural network is then used to regress the camera’s rotation deviation relative to its initial position. Unlike vehicle cameras, roadside cameras are typically hinge-mounted on infrastructure, limiting significant translational shifts. Minor deviations (centimeter-scale) have negligible impact on perception in V2X scenario~\cite{yang2023bevheight,scholler2019targetless}, whereas small rotational shifts, given the wider field of view, greatly degrade image quality (see Sec.~\ref{sec:4.5}). Thus, our primary focus is on the rotational deviation relative to the initial setup.

Inspired by the optical flow estimation~\cite{teed2020raft,wang2025sea} and lidar-camera calibration task~\cite{liao2023deep,lv2021lccnet}, we project vehicle LiDAR point clouds onto roadside camera images to generate a depth map. Then, using a ResNet + FPN (Feature Pyramid Networks)~\cite{lin2017feature}, we extract multi-scale features to form a 4D correlation volume. We iteratively update a $2\times H \times W$ feature matrix representing pixel-level correspondences between the depth map and the RGB image, using this matrix to associate new features from the 4D correlation volume. Our Mamba architecture further integrates spatial and temporal data, combining time-series information with feature matrices from prior iterations to overcome single-frame perception limitations. We divide the training process into two stages which will be explained in Sec.~\ref{sec:training}.

Compared to traditional roadside camera calibration methods, our approach fully utilizes environmental data, including but not limited to the vehicle itself. Meanwhile, it allows calibration at any road location as needed, without any pre-prepared reference equipment or images for these cameras, making it highly deployable. Compared to the existing LiDAR-camera calibration algorithms used on a single vehicle, our carefully designed network mitigates the issue of point clouds collected by vehicles in different frames occupying a relatively small and uncertain area in the roadside camera’s view, delivering robust calibration results in V2X scenarios with fewer parameters.

Our main contributions are summarized as follows: 
\begin{itemize}
    \item To the best of our knowledge, we are the first to propose a target-less V2X-based roadside camera calibration solution. 
    \item We designed a novel calibration network that combines multi-scale processing, iterative updates, and the mamba architecture to address the instability issues encountered when directly applying traditional LiDAR-camera calibration methods to V2X scenarios, while also improving calibration efficiency.
    \item We validated the effectiveness and robustness of our method on the V2X-Seq~\cite{yu2023v2x} and TUMTraf-V2X~\cite{zimmer2024tumtraf} datasets.
\end{itemize}

\section{Related Works}
\label{sec:works}
\textbf{Roadside camera calibration:\quad}
A classic approach to roadside camera calibration is target-based calibration, typically using checkerboard patterns or other reference objects~\cite{zhang2000flexible}. However, this method demands extensive manual effort and is impractical in busy traffic scenarios. To address this, some studies leverage vehicle-provided positioning or motion trajectories, aligning them with detection boxes~\cite{tsaregorodtsev2023automated,sochor2017traffic} or estimated trajectories~\cite{zhao2024graph} from roadside cameras. While eliminating the need for placing specific targets, these approaches depend heavily on detection accuracy, leading to error accumulation. Other methods exploit road geometry, such as vanishing points~\cite{kocur2021traffic}, or match perspective features like parallel curves~\cite{corral2014automatic}, but suffer from limited accuracy in complex scenes. More recent work utilizes prepared reference images or 3D reconstructions of road environments~\cite{d2024automated,vuong2024toward,ataer2014calibration}, which are labor-intensive to collect and unsuitable for real-time deployment. Co-CalibNet~\cite{yaqing2025robust} introduces additional roadside sensors, but this increases roadside deployment costs and calibration workload. To overcome these limitations, we propose a fundamentally different, target-less calibration framework that utilizes vehicle sensor data to assist roadside camera calibration without relying on physical targets, vision-based detection, or pre-built scene representations.

\textbf{LiDAR-camera calibration:\quad}
LiDAR-camera calibration methods can be broadly categorized into target-based and target-less methods~\cite{liao2023deep}. Traditional methods typically rely on specific calibration targets, such as checkerboards, spherical markers, or custom boards, transforming the calibration problem into an optimization problem. Recent research has shifted towards target-less methods, primarily applied to extrinsic parameter estimation for monocular camera and LiDAR systems on a single car. \citet{li2023automatic} categorizes target-less calibration methods into four types: information-theoretic, feature-based, ego-motion-based, and learning-based approaches. Among these, deep learning methods have garnered widespread attention due to their promising results. RegNet~\cite{schneider2017regnet} was the first to adopt a deep learning-based approach, using a network to extract features and then regress the parameters. CalibNet~\cite{iyer2018calibnet} introduced a geometry-supervised deep network to perform real-time calibration by maximizing geometric and photometric consistency. LCCNet~\cite{lv2021lccnet} and CaLiCaNet~\cite{rachman2023end} utilized cost volume to establish correspondences between features, which significantly improved upon previous deep learning-based calibration methods. \citet{zhu2023calibdepth} use depth estimation to mitigate the modal differences. The latest researches~\cite{xiao2024calibformer,lee2024lccraft} attempt to enhance the form of cost volume using new architectures like Transformer. However, these deep learning-based methods perform poorly in V2X scenarios. That's because vehicle-mounted LiDAR often covers only a small portion, posing higher demands for cross-modal feature matching. Moreover, the covered area of depth maps remains nearly constant in the single vehicle's view, while the point clouds collected by the vehicle occupy different regions over time from a roadside perspective. Our method aims to address these problems by using temporal information and a better cost volume update mechanism.

\textbf{State Space Models:\quad}
Recently, State-Space Models (SSMs) have excelled in modeling long sequences, particularly in capturing long-range dependencies. The Structured State-Space Sequence (S4)~\cite{gu2021efficiently} has been proposed as an alternative to Convolutional Neural Networks (CNNs) and Transformers, offering the advantage of linear complexity for efficient handling of long sequences. As research on S4 continues to deepen, various improved models have emerged, such as the S5 layer~\cite{smith2022simplified} which introduces MIMO SSMs and efficient parallel scanning and the Gated State-Space layer (GSS)~\cite{mehta2022long}. These models have shown promising performance in language modeling tasks. Excitingly, the recent introduction of the data-dependent SSM layer has led to the development of Mamba~\cite{gu2023mamba}, which outperforms Transformers of various sizes on large-scale real-world data and achieves linear scaling in sequence length. Recent studies have attempted to bring Mamba model into the visual domain. Vision Mamba~\cite{zhu2024vision} borrows the concept from Vision Transformers~\cite{liu2021swin} by marking image sequences with position embeddings and compressing the visual representation using bidirectional state space models. Furthermore, Video Mamba~\cite{li2025videomamba} offers a linear-complexity approach for modeling dynamic spatiotemporal contexts, making it ideal for high-resolution long videos. Research indicates that the Mamba architecture has advantages over LSTM and Transformer architectures in video processing tasks. Building on Mamba, our work aims to achieve more efficient temporal modeling with reduced computational cost. We draw on visual-domain Mamba applications to integrate LiDAR and roadside camera temporal features at the representation level, thereby mitigating calibration failures caused by poor single-frame data quality.
\section{Method}
\begin{figure*}[t]
  \centering
  \includegraphics[width=0.9\linewidth]{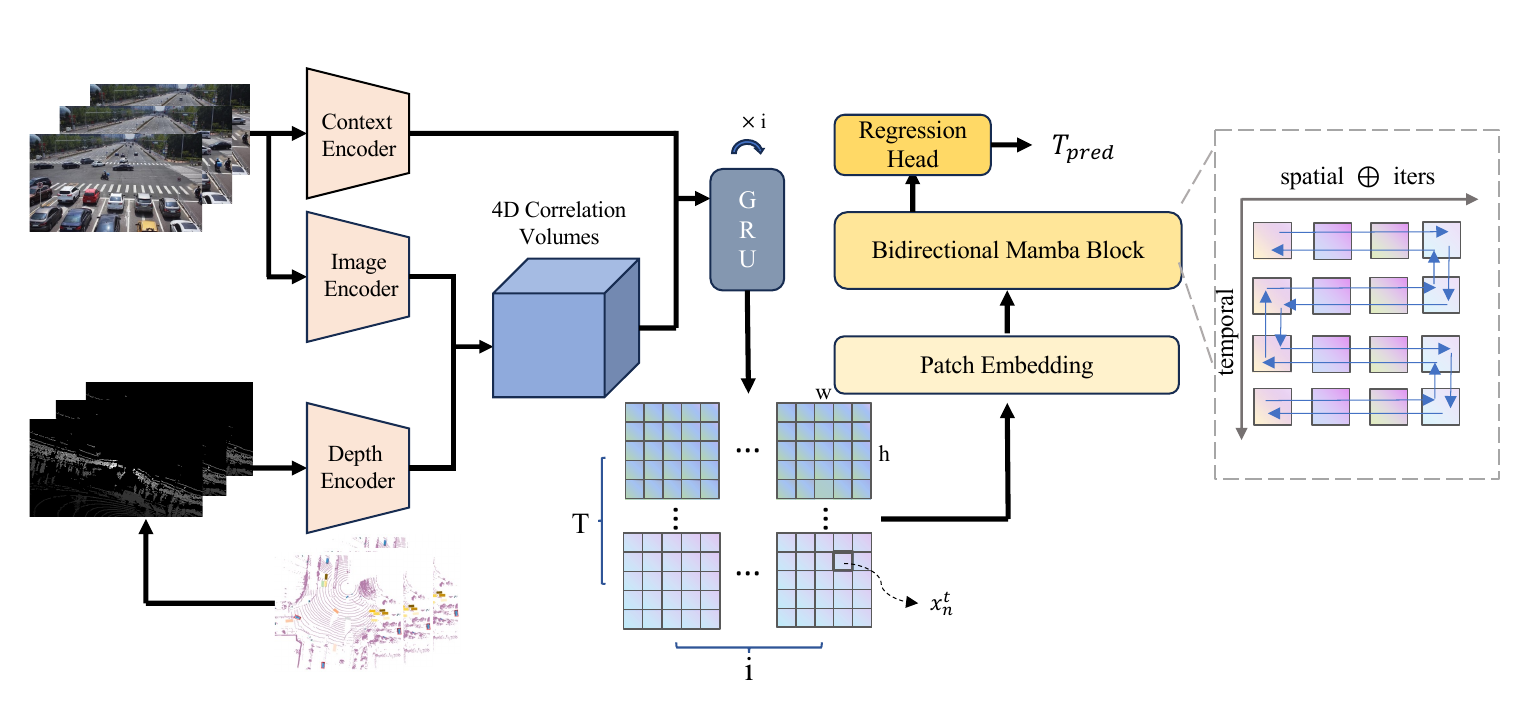}
  
  \caption{The overall framework of MamV2XCalib. Initially, the point clouds are projected into the camera view using inaccurate initial extrinsic parameters. The network input consists of multiple frames of erroneous depth maps and roadside camera images. Multi-scale features are extracted from the images and depth maps through three channels, forming 4D correlation volumes. A GRU block is utilized to recurrently update the pixel-level correspondences between the images and depth maps (calibration flow) by retrieving features from the volume based on existing estimates. A series of calibration flows generated from multiple frames and iterations is divided into patches and fed into the Mamba model, with temporal and spatial embeddings. An additional token summarizes the features to regress the rotation deviation value.}
  \label{fig:main}
\end{figure*}

In this section, we present MamV2XCalib, a novel roadside camera calibration method. The framework of our approach is shown in Fig.~\ref{fig:main}. In the first stage, we review the mathematical definition of the roadside camera calibration problem (Sec.~\ref{sec:3.1}) so as to better understand our newly proposed calibration strategy. Next, we provide a detailed explanation of our network design, loss function, and inference process.
\subsection{Problem Formulation and Input Processing}
\label{sec:3.1}
The modeling of a standard perspective camera can be described by the correspondence between 3D spatial points and pixels.
\begin{equation}
d_i 
\begin{bmatrix}
u_i \\
v_i \\
1
\end{bmatrix} 
= \mathbf{K} \mathbf{T}
\begin{bmatrix}
x_i \\
y_i \\
z_i \\
1
\end{bmatrix} 
= \mathbf{K} 
\begin{bmatrix}
\mathbf{R} & \mathbf{t} \\
\mathbf{0} & 1
\end{bmatrix} 
\begin{bmatrix}
x_i \\
y_i \\
z_i \\
1
\end{bmatrix}
\label{proj}
\end{equation}
where$\begin{bmatrix}
x_i & y_i & z_i &1
\end{bmatrix}^T$represents the homogeneous coordinates of a 3D point in an arbitrary known coordinate system $A$ and $\begin{bmatrix}
u_i & v_i &1
\end{bmatrix}^T$ denotes a homogeneous camera image coordinates of a picture. $\mathbf{K}$ denotes the camera’s intrinsic parameters, while $\mathbf{R}$ and $\mathbf{t}$ represent the rotation and translation matrices, which transform the coordinates from system $A$ to the camera coordinate system. $d_i$ represents the projection depth of the 3D point. Specifically, in our method, $A$ is set as the vehicle LiDAR coordinate system. Correspondingly, $\mathbf{R}$ and $\mathbf{t}$ represent the transformation from the onboard lidar coordinate system to the camera coordinate system. According to the description in the introduction, we assume that these parameters’ initial values are given according to the expected calibration values, denoted as $K_{init}$, $R_{init}$, $t_{init}$, where $K_{init}$ and $t_{init}$ are stable and easy to deal with, and changes mainly occur in $R_{init}$:
\begin{equation}
T_{init}=\Delta T \cdot T_{LC}
\end{equation}
\begin{equation}
\Delta T= \begin{bmatrix}
\mathbf{R_{error}} & \mathbf{0} \\
\mathbf{0} & 1
\end{bmatrix} 
\end{equation}
where $T_{LC}$ represents the transformation from the current LiDAR coordinate system to the camera coordinate system. $\mathbf{R_{error}}$ is the rotational deviation. $T_{init}$ is artificially set and our goal is to automatically obtain the value of $\Delta T$ through sensor data, which can determine the current position of the roadside camera in the world coordinate system. Before being fed into the network, the sensor data needs to be aligned to a unified coordinate system. The vehicle’s point cloud is projected onto the roadside camera plane using the initial parameters $T_{init}$ and Eq.~\ref{proj}.
\subsection{Multi-scale feature extraction}
Our method employs three feature extraction branches to separately extract image information, depth map information, and image context. The two branches processing camera images both use a pre-trained ResNet-18 as the backbone. To obtain richer multi-scale features, and considering that highly abstract feature maps are not conducive to the subsequent search for pixel-level relationships, we fuse the upsampled feature map of latter layers with the former layers to obtain the final feature map. Finally, the output of image encoder $F_1$ and $F_{context}$ is $\mathbb{R}^{\frac{H}{8} \times \frac{W}{8} \times D}$ where we set $D=256$. The depth map branch architecture $F'$ is similar to the image branch, except that the input channel is modified to one dimension, and pre-trained parameters are not loaded. Due to the unique nature of depth maps, the three branches do not share parameters.

\subsection{Feature Matching and Iterative refinement}\label{sec:3.3}
We directly associate the correspondences between roadside images $I_1 \in \mathcal{I}$ and depth maps $I_2 \in \mathcal{D}$ with the deviation of the roadside camera’s extrinsic parameters. This pixel-level correspondence can be regarded as a special type of flow field, which we refer to as the \textit{calibration flow} in this paper. Inspired by the RAFT~\cite{teed2020raft} model used for optical flow estimation, we construct a multi-scale 4D correlation volume as follows: 
\begin{equation}
Inn(F_1(I_1), F'(I_2)) \in \mathbb{R}^{h \times w \times h \times w}
\end{equation}
\begin{equation}
U_k=\text{Pool}(Inn) \in \mathbb{R}^{h \times w \times \frac{h}{2^k} \times \frac{w}{2^k}}
\end{equation}
where $Inn$ is calculated through inner product and Pool is designed to get multi-scale 4D correlation volume. Through these operations, the similarity between any two pixels is stored in this volume. Next, we repeatedly query features from the 4D dictionary to estimate pixel correspondences  based on the former correspondences. We initialize the flow field (pixel correspondences) to zeros. In each iteration, given the feature flow map, we take each pixel’s corresponding point as the center and get the similarity of all pixels within a distance of $r$. We call this operation \text{LookUp}.
\begin{align}
    h''&= \text{GRU}(\text{LookUp}\{U_k,f,r\},h',F_{context}(I_1))\\
    df &= \text{Flow}(h'')
\end{align}
We then input the features we just looked up, hidden states from last iteration, and context features together into the GRU module, obtaining new hidden states. The flow residuals $df$ is predicted by layer Flow consisting of two convolutional layers. The residual $df$ is added to the previous flow field to update it. In this way, we obtain a series of flow field estimates $f_i \in \mathcal{F}$ equal to the number of iterations.

To ensure that the model can get calibration flow fields, in the first stage of training, we directly regress the extrinsic parameters from each flow field obtained at every step using two fully connected layers.

\begin{table*}
\centering

\begin{tabular}{llcccccccc}
\toprule
\multicolumn{2}{c}{} & \multicolumn{4}{c}{\textbf{Mean ($^\circ$)}} & \multicolumn{4}{c}{\textbf{Std ($^\circ$)}} \\
\cmidrule(r){3-6} \cmidrule(l){7-10}
\multicolumn{1}{c}{\textbf{Deviation}} & \multicolumn{1}{c}{\textbf{Method}} & Total & Roll & Pitch & Yaw & Total & Roll & Pitch & Yaw \\
\midrule

$(-20^\circ, +20^\circ)$ &LCCNet~\cite{lv2021lccnet} & 1.0741 &\bf{0.1788} &0.6056 & 0.6455  & 2.5336 &0.1537 &2.4624& 0.7920  \\
$(-20^\circ, +20^\circ)$ &Calib-anything~\cite{luo2023calib} & - &- &- & -  & - &- &-& -  \\

$(-20^\circ, +20^\circ)$ &Ours & \bf{0.6313} & 0.1867 & \bf{0.2409} & \bf{0.4780} & \bf{0.3211} & \bf{0.1489} &\bf{0.1860} &\bf{0.3500}  \\
\midrule
$(-10^\circ, +10^\circ)$ &LCCNet~\cite{lv2021lccnet} &0.7909&\bf{0.1792} &0.3440 & 0.5548 &1.6825  & \bf{0.1111}&1.6147 & 0.5985  \\
$(-10^\circ, +10^\circ)$ &Calib-anything~\cite{luo2023calib} & - &- &- & -  & - &- &-& -  \\

$(-10^\circ, +10^\circ)$ &Ours & \bf{0.5858} &0.2260   & \bf{0.2441} & \bf{0.4001 }& \bf{0.2428} & 0.1525 & \bf{0.1745}& \bf{0.2776} \\
\bottomrule

\end{tabular}
\caption{Comparison results with other calibration methods on V2X-Seq dataset~\cite{yu2023v2x}. Calib-anything~\cite{luo2023calib} fails. (See Fig.~\ref{fig:calibany})}
\label{tab:my_table}
\end{table*}

\subsection{Global Aggregation with Mamba}
While our 4D correlation volume effectively captures global correlations between inaccurate depth maps and images, the unique V2X scenario poses additional challenges. The already sparse point clouds become even sparser due to viewpoint disparities between roadside and vehicle perspectives. Moreover, the effective regions of depth maps shift over time, making it inherently difficult to derive accurate calibration flow fields in uncertain, data-scarce regions using single-frame data alone. Directly regressing extrinsics from all correspondences would inevitably introduce bias.

Fortunately, although the vehicle is in motion, the roadside camera’s deviation from its initial position remains fixed over a calibration period. This stability allows us to perform temporal fusion, integrating data across frames to overcome the limitations described above. We select Mamba because our sequence of calibration flow fields has dimensions akin to video data, where Transformers excel in fusion tasks. However, our iterative process produces a large number of tokens, prompting us to adopt the latest Mamba~\cite{li2025videomamba,zhu2024vision} for its superior efficiency in handling such data.

Specifically, we extend the input from a single frame to a sequence. According to the method in Sec.~\ref{sec:3.3}, we can obtain a calibration flow field of size $\mathbb{R}^{T \times i \times h \times w}$. We treat the iterative process and temporal dimension as two separate axes. First, we concatenate the flow field maps obtained from each frame into an overall map of size $\mathbb{R}^{T \times h \times (w\times i )}$. Then  we divide it into a series of patches $x_n^t$ instead of flattening them simply and project them into a hidden space through convolutional layers. The patches of the frames are arranged as shown in Fig.~\ref{fig:main}. Similar to \cite{arnab2021vivit,li2025videomamba}, we add positional and temporal embeddings $p_s$ and $p_t$ to retain the spatiotemporal distinctions and insert an additional token $z_{add}$ into the entire sequence to summarize the information.
\begin{align}
    Z&=[z_{add},\mathbf{E}x_1^1,\mathbf{E}x_2^1,\ldots,\mathbf{E}x_N^T]+p_s+p_t\\
    z_{add}&=\mathbf{Mam}(Z)
\end{align}
where the projection by $\mathbf{E}$ is equivalent to 3D convolution and $\mathbf{Mam}$ is the bidirectional mamba block. Through the Mamba model, we construct state space and scan it as illustrated, enabling the model to selectively model different regions of each flow field map from a spatial perspective. Additionally, this allows the model to automatically associate features generated from data with varying quality at different timestamps. The added token summarizes the series of lengthy calibration flow maps generated through iterations and temporal processes.

After obtaining the refined summarized features $z_{add}$, we use two fully connected layers to regress the quaternion, achieving the desired estimation of the rotational error.

\subsection{Loss Function}
For each obtained rotation prediction, our loss has the following form:
\begin{equation}
L_i=\lambda_r L_r +\lambda_p L_p
\label{eq:loss}
\end{equation}
where $\lambda_r$ and $\lambda_p$ denote respective loss weights.
$L_r$ is the rotation loss and we use angular distance to represent  the difference between quaternions. Considering that our method utilizes coordinate transformations of point clouds $P_i$, we calculate the distance between the two point clouds after the actual transformation and the predicted transformation to obtain $L_p$:
\begin{equation}
L_p = \frac{1}{N} \sum_{i=1}^{N} \left\| T_{LC}^{-1} \cdot T_{pred}^{-1} \cdot T_{init} \cdot P_i - P_i \right\|_2
\end{equation}
where $T_{LC}$ is the ground truth LiDAR-Camera extrinsic matrix and $T_{pred}$ is the output of the calibration network.

In the first training stage, we iteratively obtain $n$ estimates per frame, with the loss as a weighted sum of their individual losses using exponentially increasing factors.
\begin{equation}
L=\sum_{i=1}^{n}\gamma ^{n-i}L_i
\end{equation}
In the second stage, multi-frame data is input, with each frame producing a single estimate, and the loss is defined as in Eq.~\ref{eq:loss}.
\subsection{Calibration Inference}
One of our inputs, the depth map, is influenced not only by the quality of the point cloud itself but also by the initially given extrinsic parameters. To address significant deviations and maintain consistency with LCCNet~\cite{lv2021lccnet}, we concatenate multiple networks that have the same architecture but are trained under different noise conditions. We then re-project the depth maps based on the predictions $T_i$ from the previous network to obtain higher-quality input data. Therefore, during the final inference, we have:
\begin{equation}
\tilde{T}_{LC} = (T_0 \cdot T_1 \ldots T_n)^{-1}T_{init}
\end{equation}

\begin{figure*}[htbp]
    \centering
    \begin{subfigure}[t]{0.24\textwidth}
        \centering
        \includegraphics[width=\linewidth]{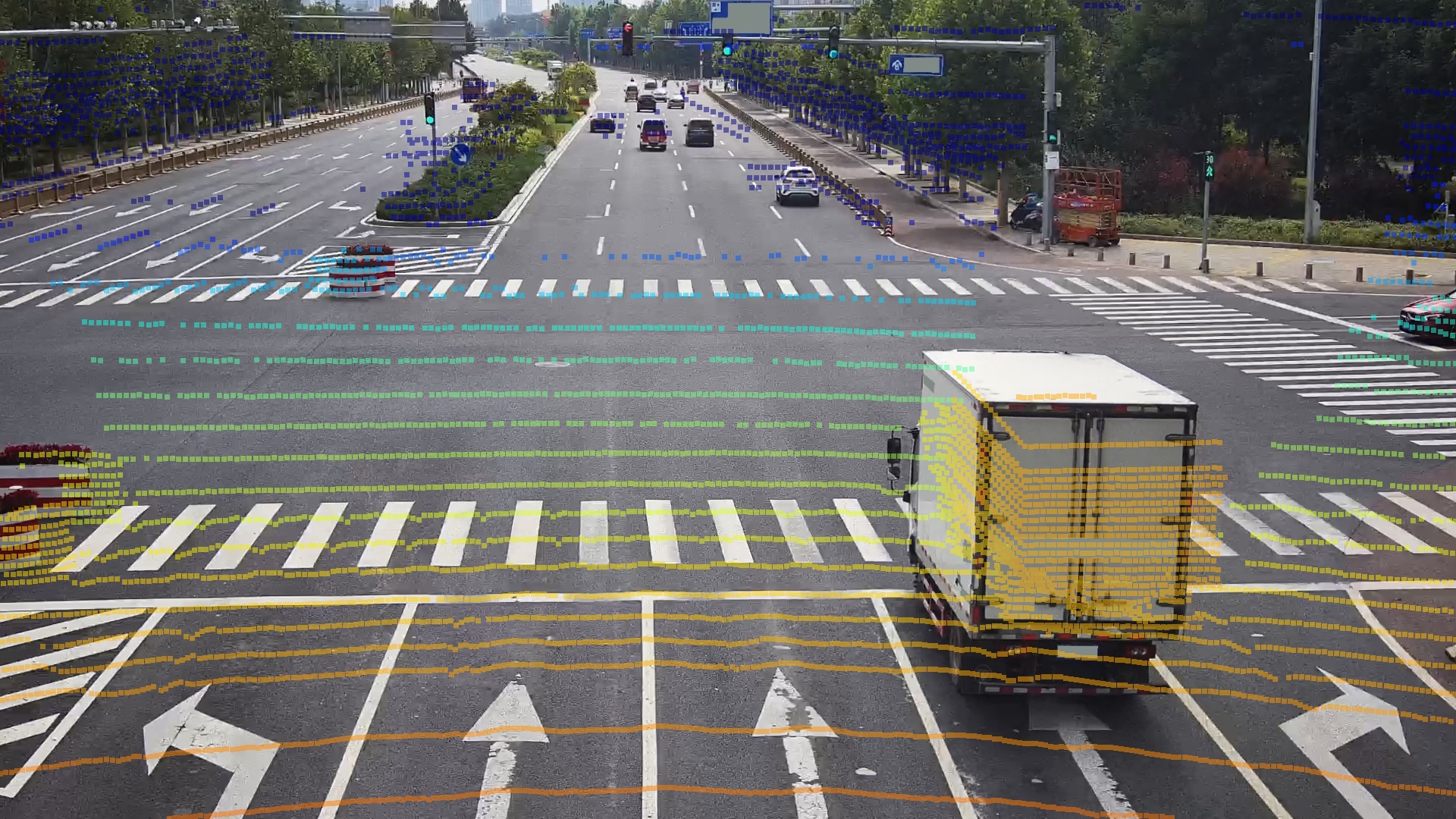}
        \includegraphics[width=\linewidth]{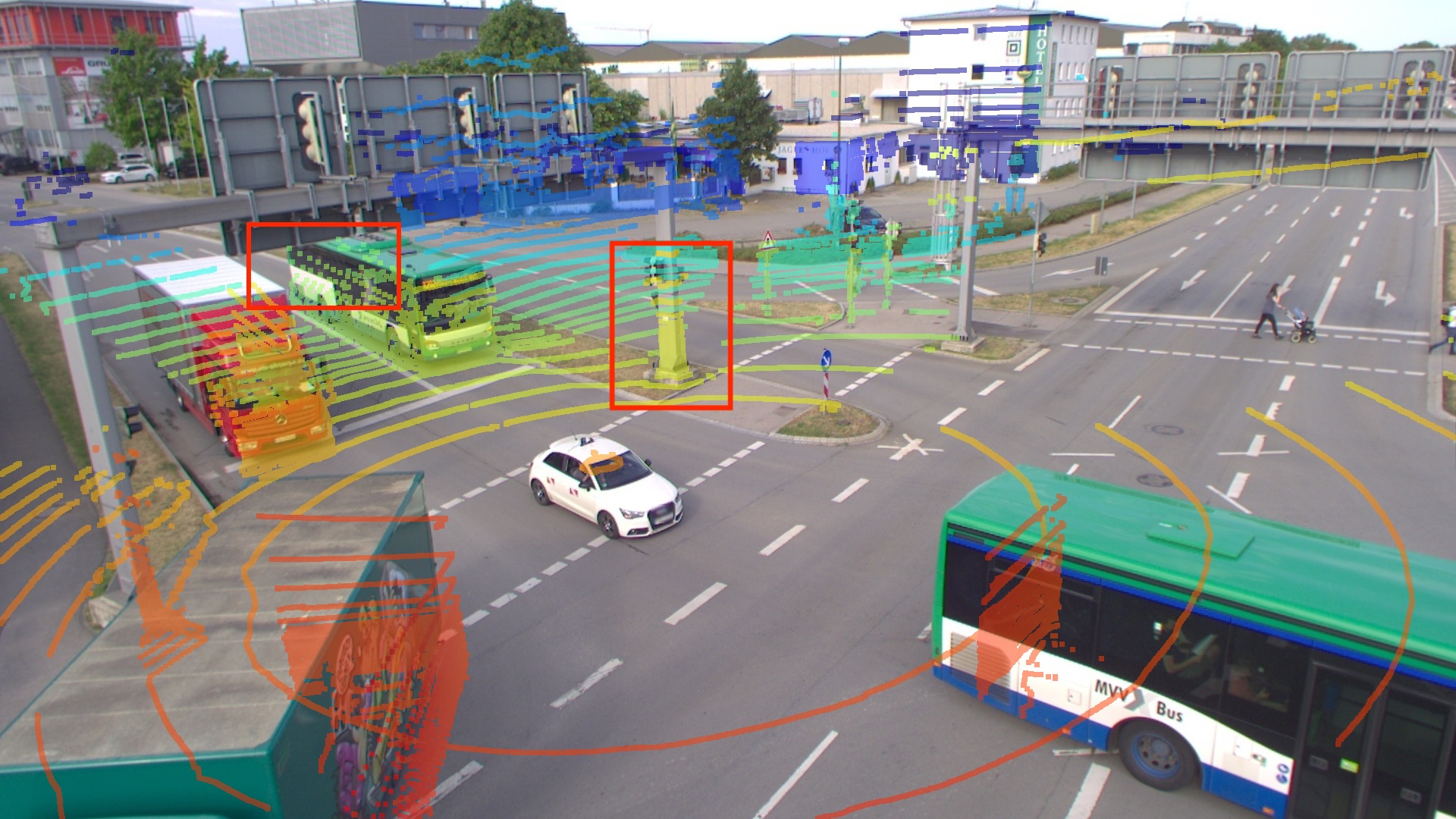}
        \includegraphics[width=\linewidth]{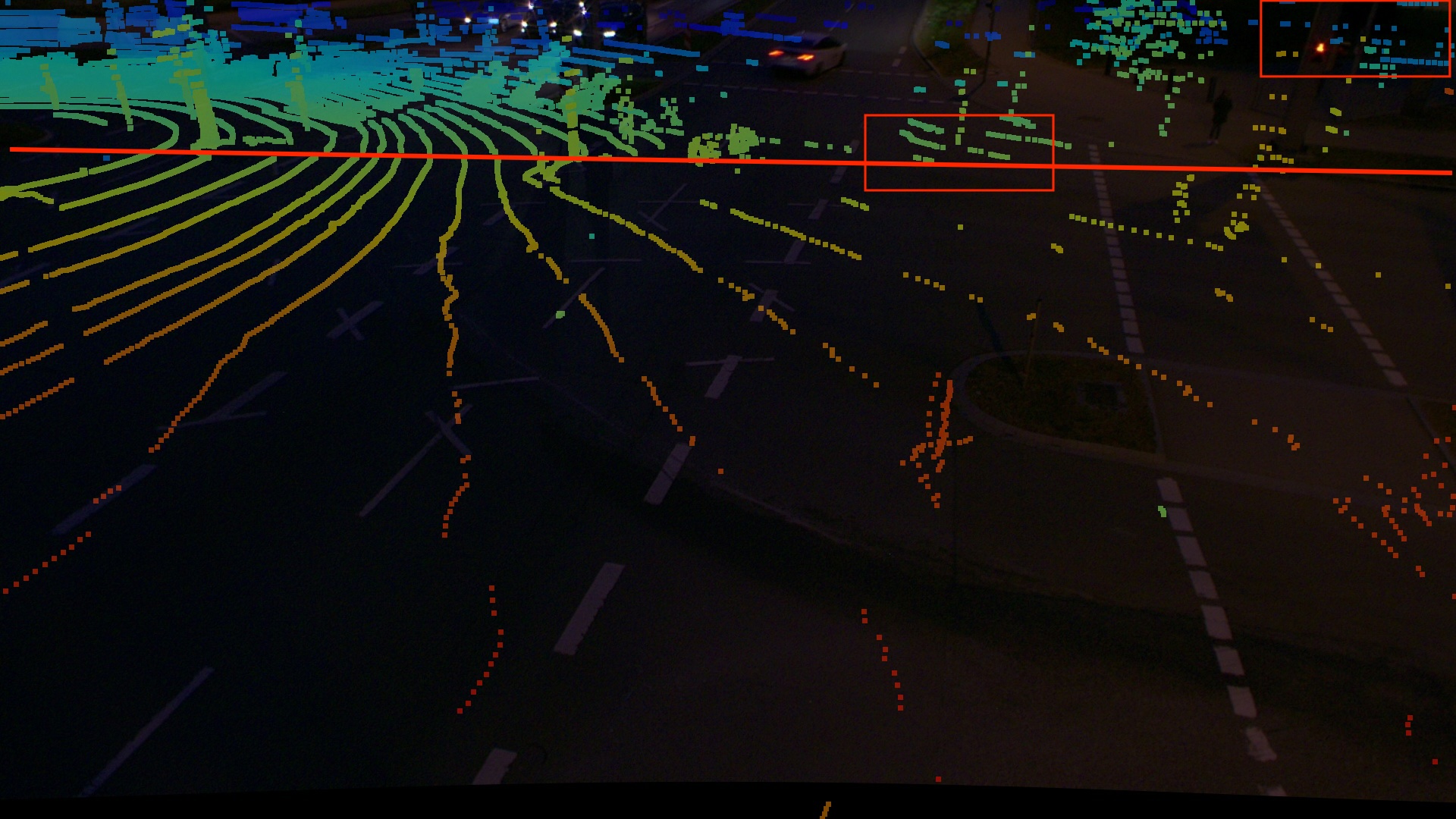}
        \caption{Ground Truth}
    \end{subfigure}
    \begin{subfigure}[t]{0.24\textwidth}
        \centering
        \includegraphics[width=\linewidth]{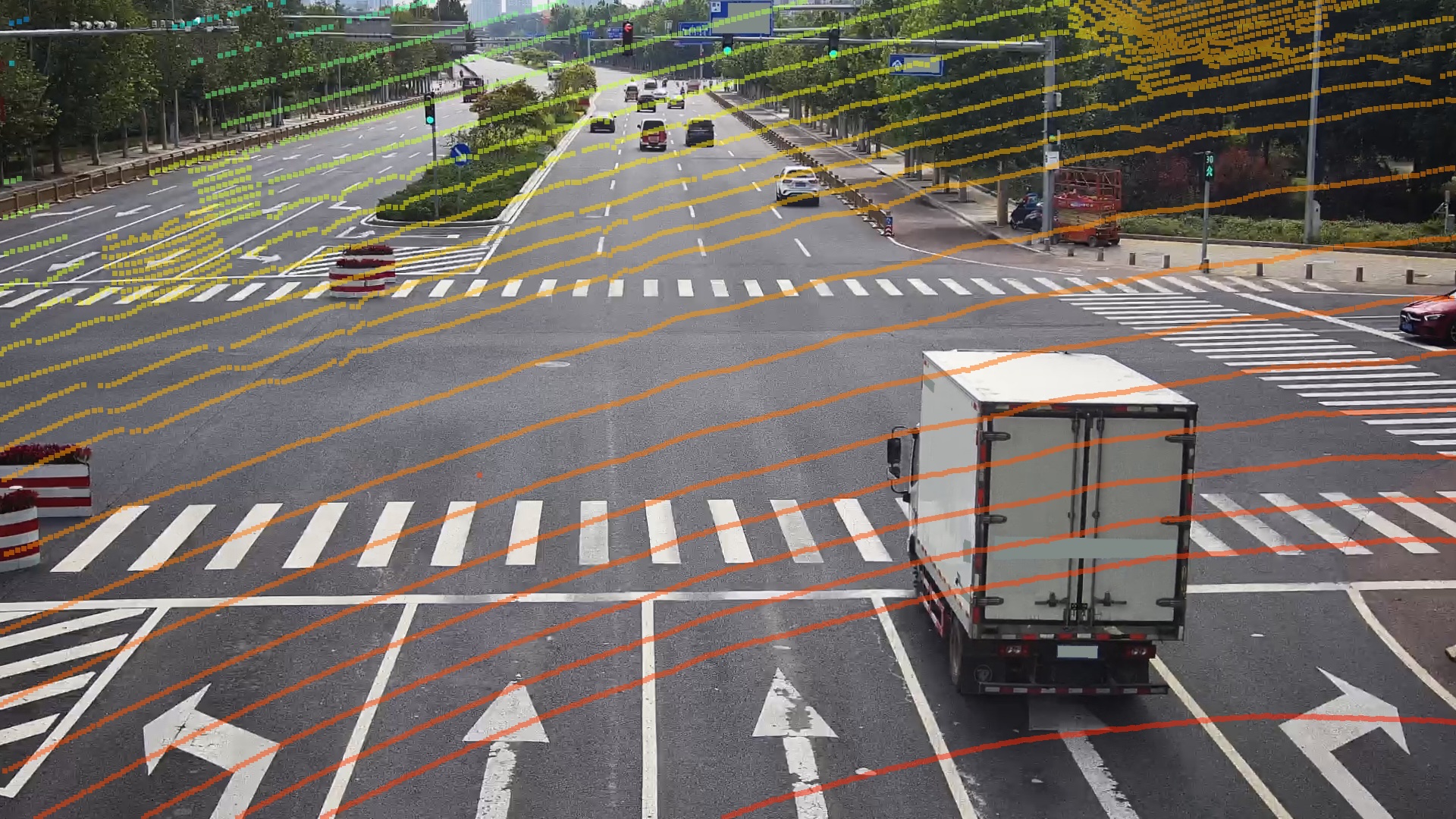}
        \includegraphics[width=\linewidth]{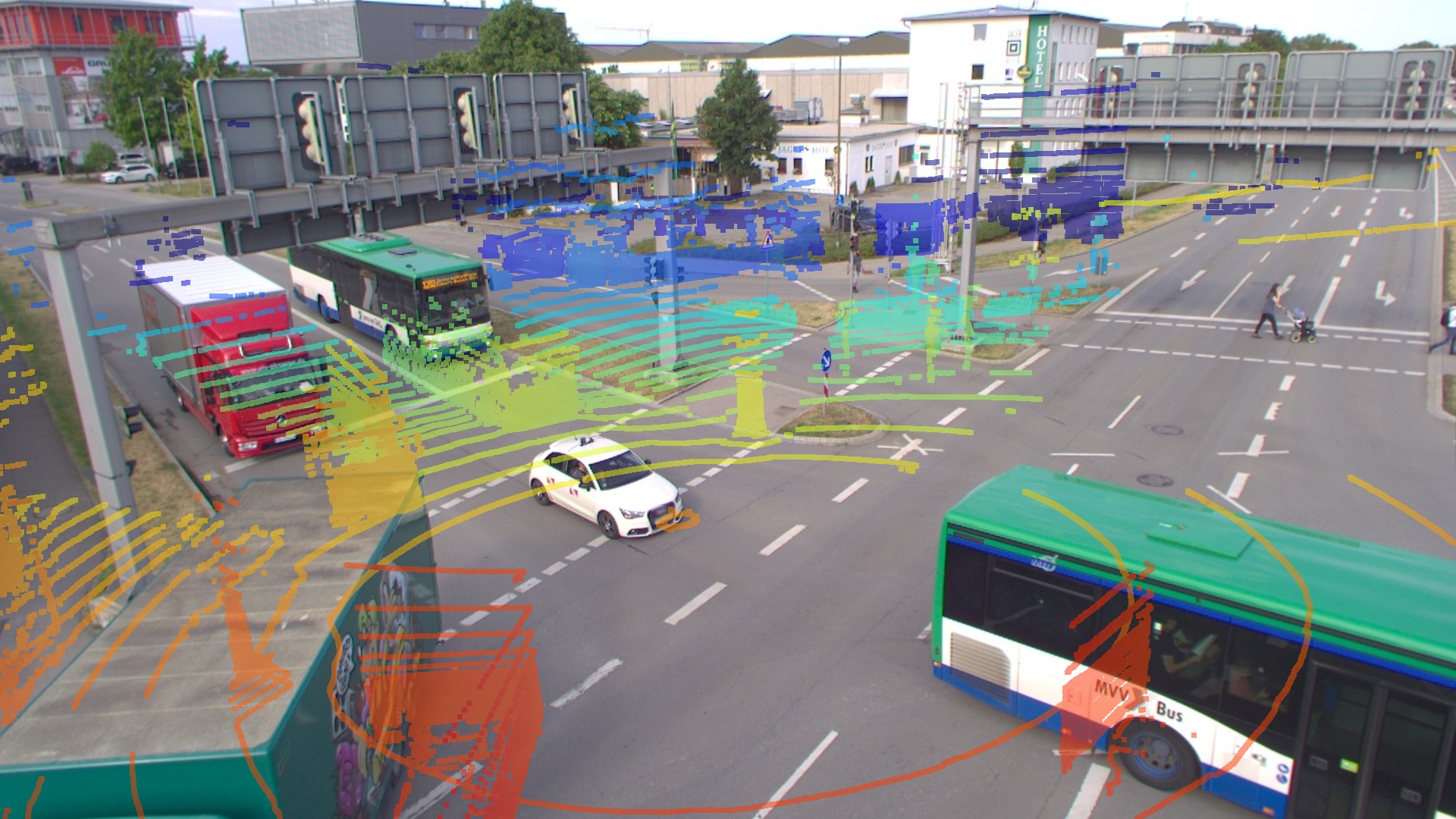}
        \includegraphics[width=\linewidth]{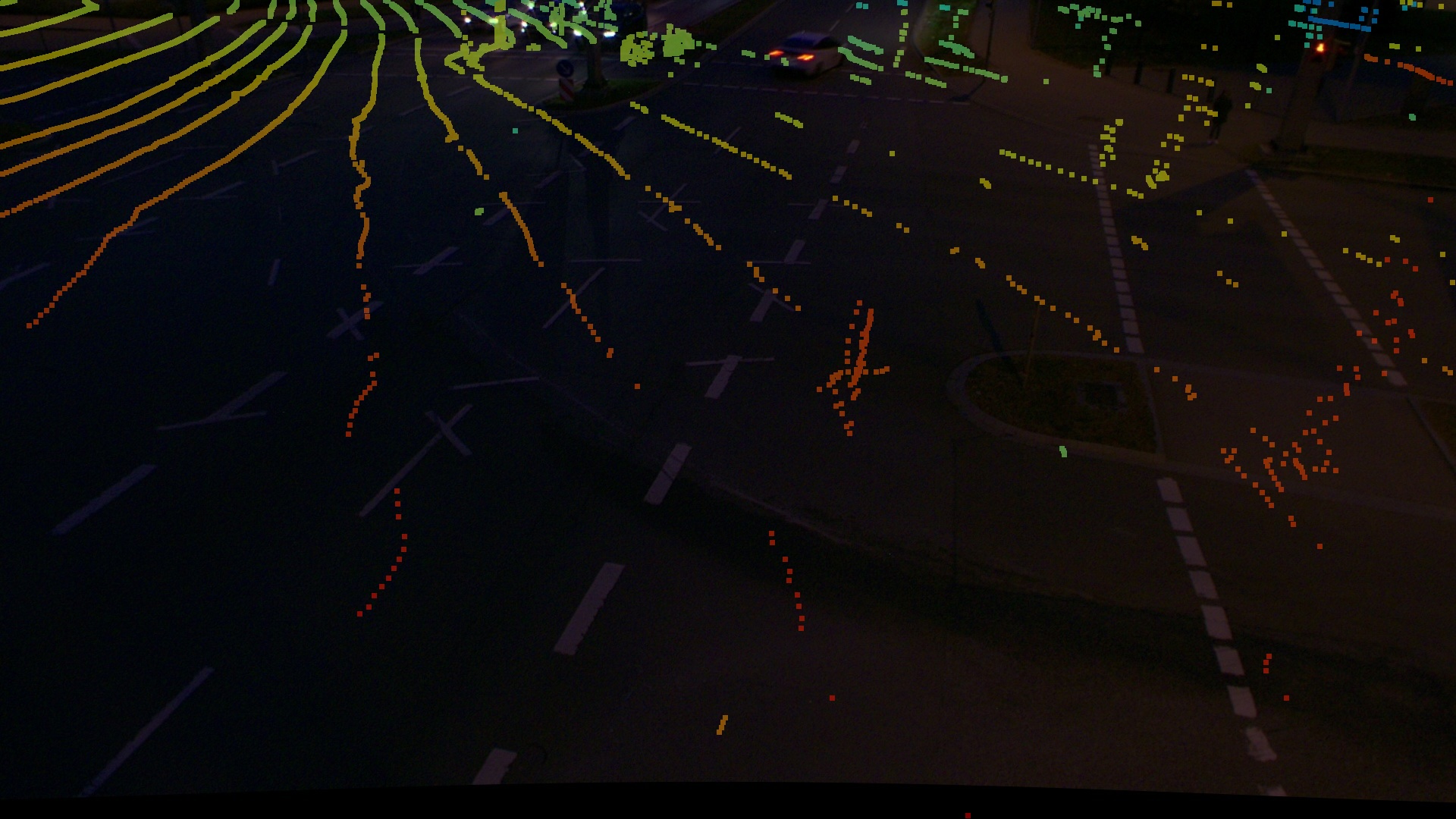}
        \caption{Before Calibration}
    \end{subfigure}
    \begin{subfigure}[t]{0.24\textwidth}
        \centering
        \includegraphics[width=\linewidth]{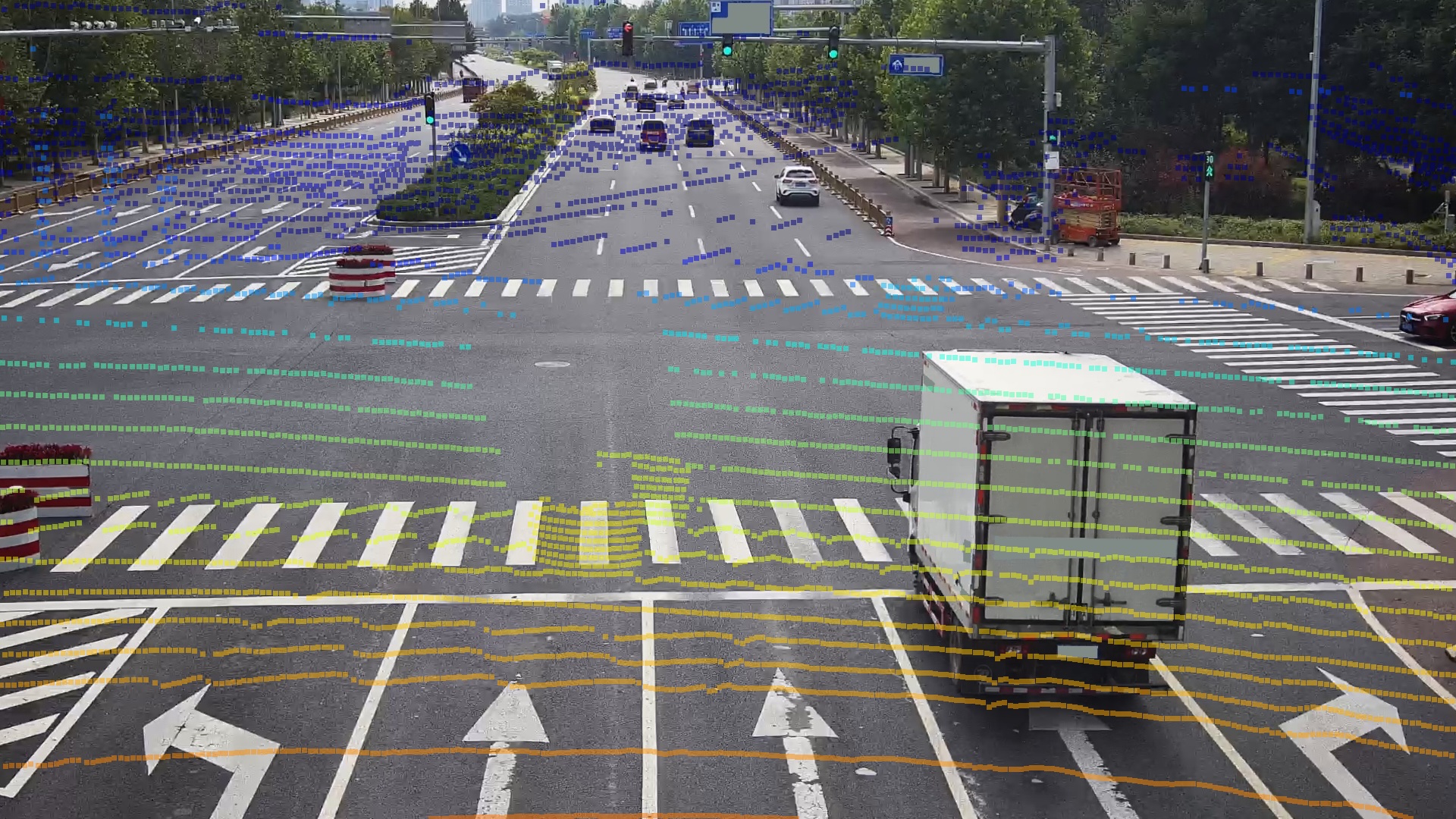}
        \includegraphics[width=\linewidth]{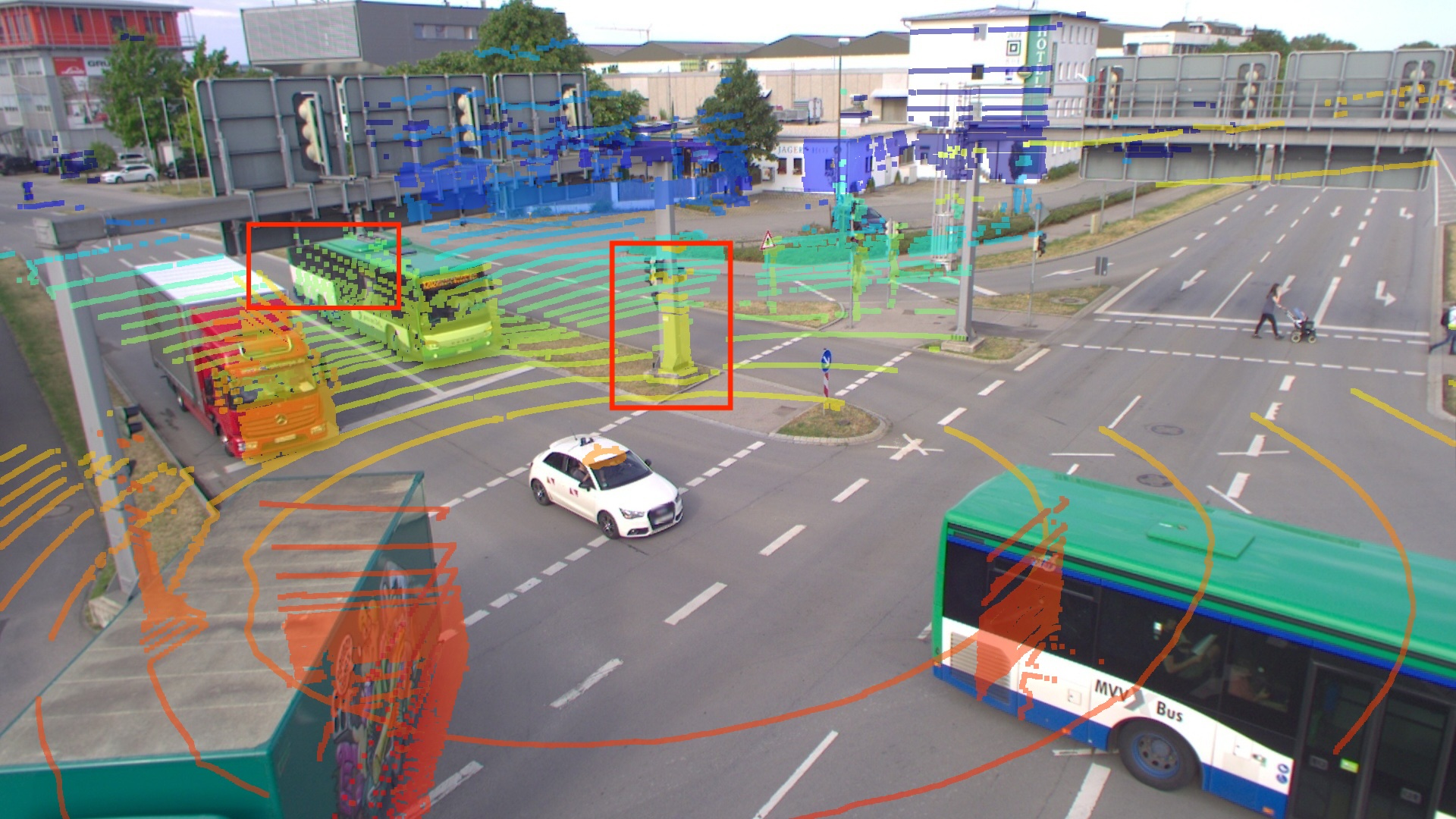}
        \includegraphics[width=\linewidth]{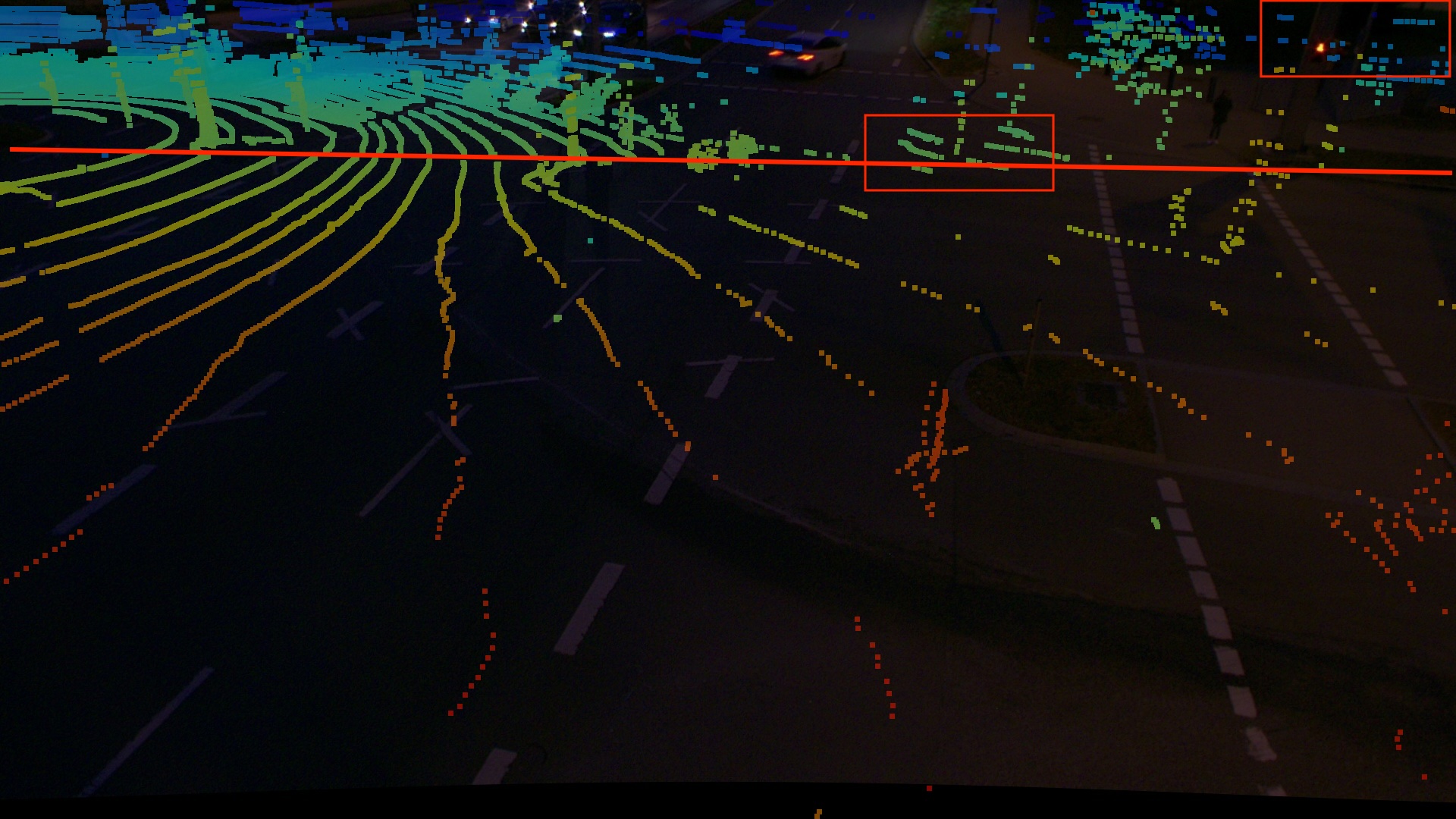}
        \caption{LCCNet}
    \end{subfigure}
    \begin{subfigure}[t]{0.24\textwidth}
        \centering
        \includegraphics[width=\linewidth]{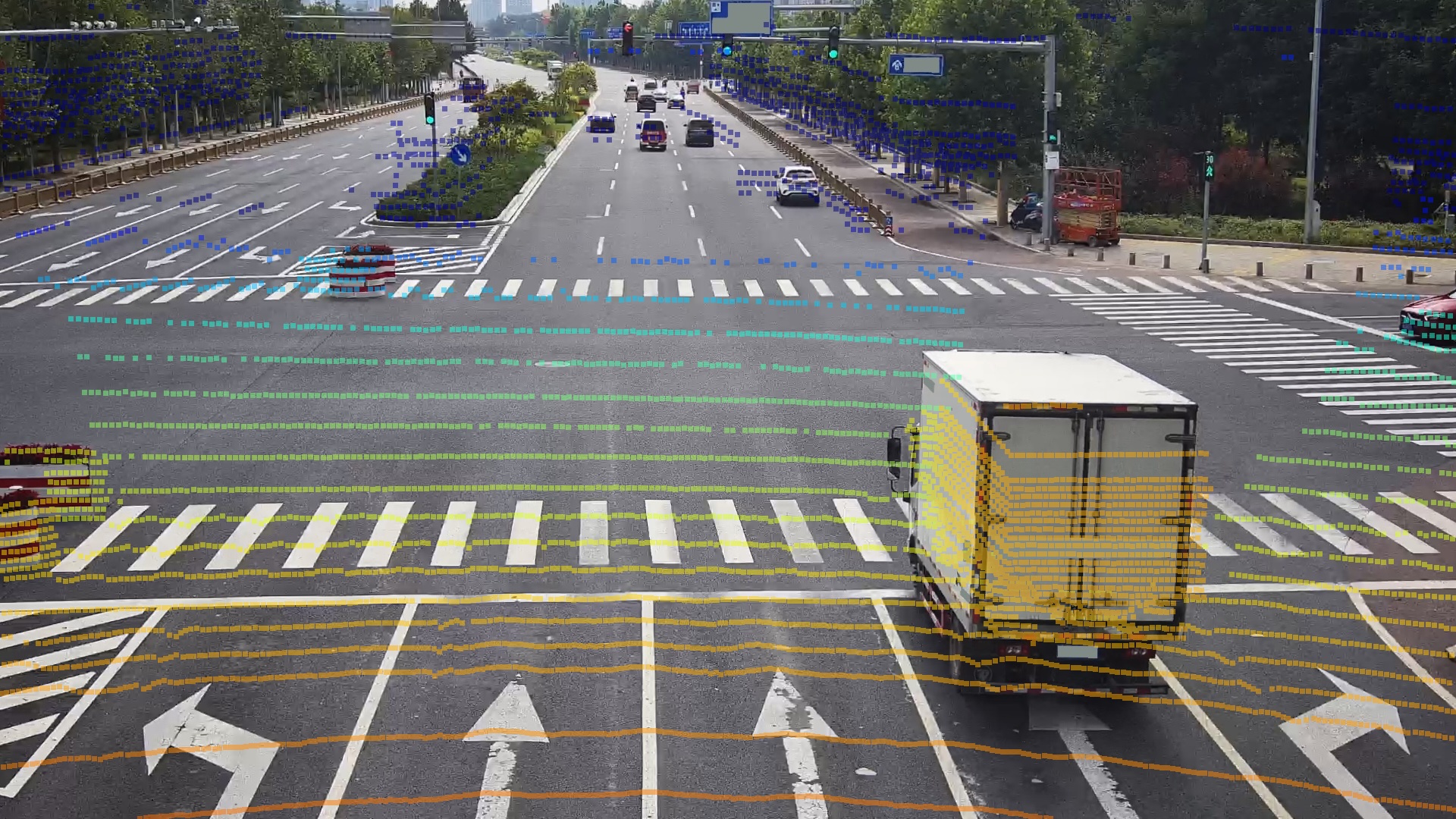}
        \includegraphics[width=\linewidth]{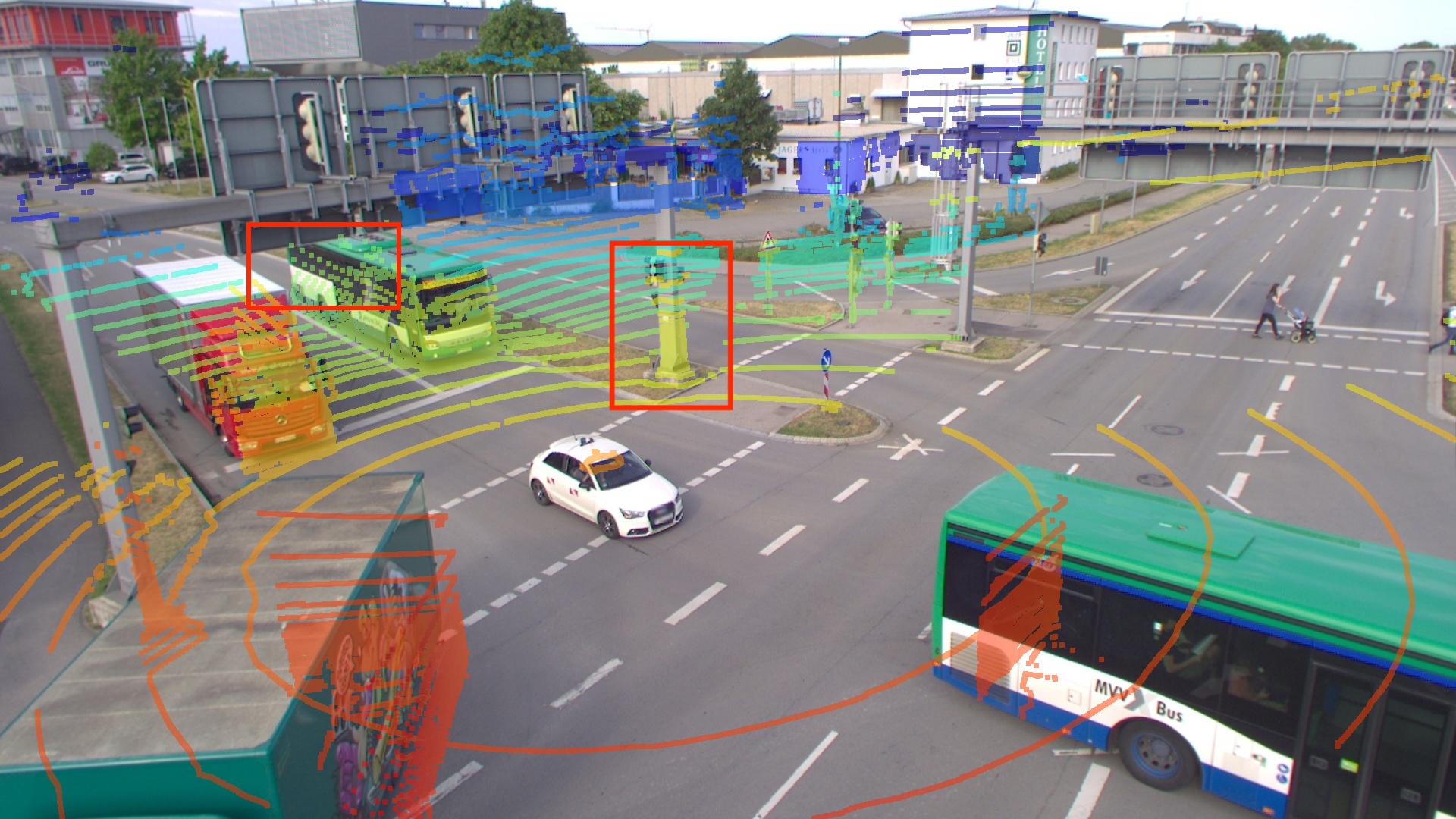}
        \includegraphics[width=\linewidth]{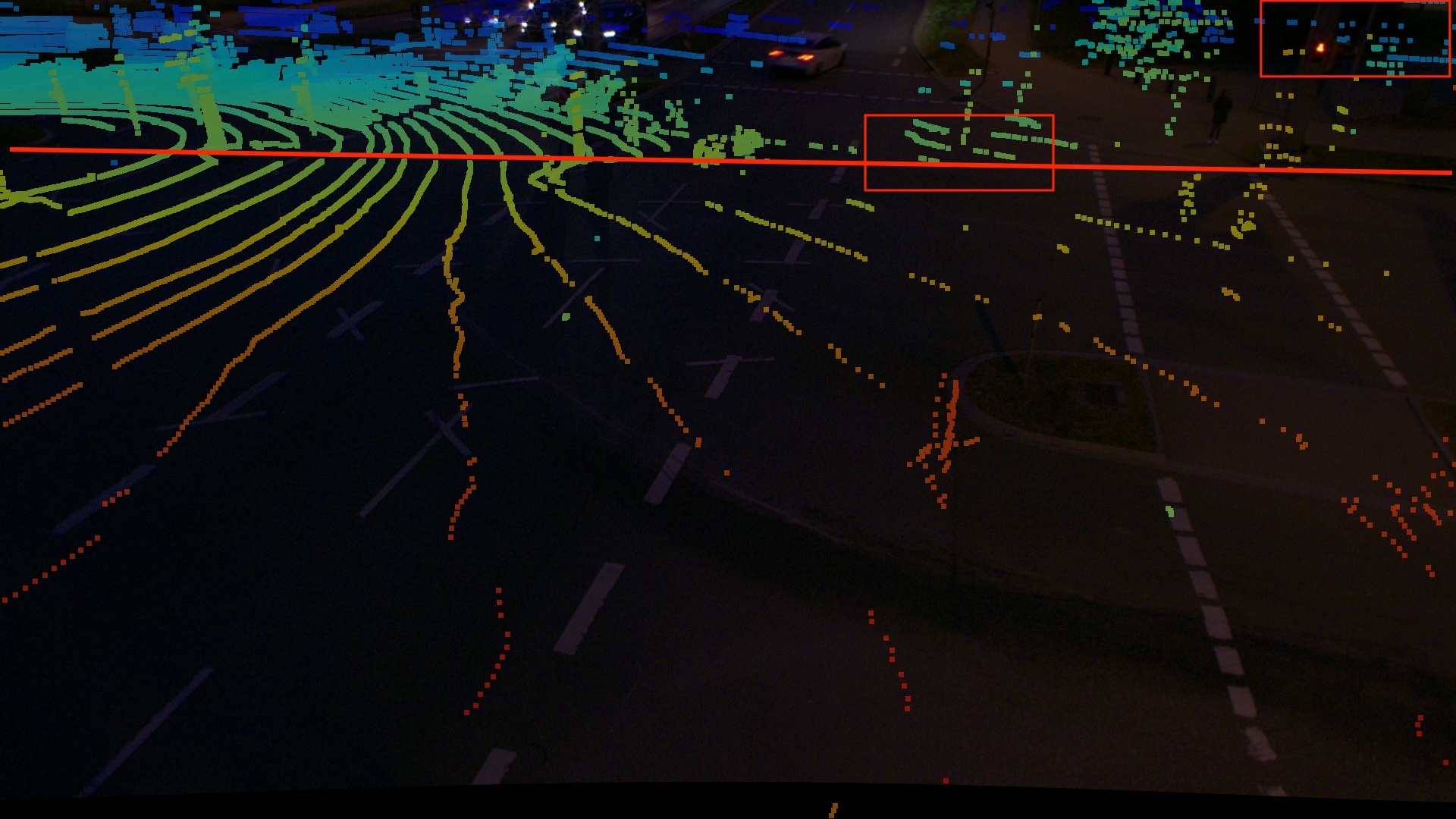}
        \caption{MamV2XCalib}
    \end{subfigure}
    \caption{Qualitative reprojection results. First row: On V2X-Seq dataset, within the initial large deflection range of $(-20^\circ, +20^\circ)$, LCCNet~\cite{lv2021lccnet} shows significant deviation in some scenes, while our method remains accurate. Second row: TUMTraf-V2X (daytime). Third row: TUMTraf-V2X (nighttime). The initial deflections are within a small range of $(-5^\circ, +5^\circ)$ and our method outperforms better.}
    \label{fig:comparison}
\end{figure*}
\section{Experiment}
\label{sec:experiment}

We conduct extensive experiments on two real-world datasets, V2X-Seq~\cite{yu2023v2x} and TUMTraf-V2X~\cite{zimmer2024tumtraf}. In this section, we will specifically introduce the training details and present the experimental results.

\begin{table*}
\centering

\begin{tabular}{lcccccc}
\toprule
Experiment & Temporal & Iteration & 4D Correlation Volume & Mean ($\circ$ )  & Std ($\circ$ ) & Param (M)  \\
\midrule
Ours  & Mamba\cite{gu2023mamba} &  \checkmark   &   \checkmark & 0.6313 &0.3211  & 47\\
w/o Mamba (+Trans)    & Timesformer\cite{bertasius2021space}  &  \checkmark   & \checkmark& 0.6683 &0.3808  & 55\\
w/o Temporal     &  -   &   \checkmark   &   \checkmark   &  0.9742  &2.0013  & 41 \\
w/o Iteration   &  Mamba\cite{gu2023mamba}   &   -   &   \checkmark   & 0.7036  & 1.6162  & 47 \\
w/o Temporal and Iteration  &  -   &   -   &   \checkmark   &  1.1025   & 2.9178  & 47 \\
3D feature matching~\cite{lv2021lccnet}    &  -   &   -   &   -   &  1.0741   & 2.5336  & 70 \\

\bottomrule
\end{tabular}
\caption{Ablation study on the effect of different components of our model.}
\label{tab:Ablation}
\end{table*}

\subsection{Dataset}

V2X-Seq: \citet{yu2023v2x} introduce a large-scale real-world cooperative vehicle-infrastructure dataset. It contains a large amount of time-synchronized data from vehicle-side and roadside sensors, as well as transformations between multiple coordinate systems.

TUMTraf-V2X: There is another V2X dataset named TUMTraf-V2X~\cite{zimmer2024tumtraf}. It focuses on challenging traffic scenarios and various day and nighttime scenes. We utilize vehicle-side LiDAR data and roadside camera data.

Similar to single-vehicle LiDAR-camera calibration tasks, we add random noise within a specified range to the extrinsic parameters to generate abundant training data. Specifically, after applying a random perturbation $\Delta T$ to $T_{LC}$, we obtain the initial extrinsic parameter $T_{init} = \Delta T \cdot T_{LC}$. By randomly changing the deviation value, we can acquire a large amount of training data.

\subsection{Training Details}
\label{sec:training}
We train our network on a single A100 GPU, with the training process divided into two stages. Next, I will elaborate on the training process and use the hyperparameter settings on V2X-Seq~\cite{yu2023v2x} as an example.

In the first stage, we perform calibration using only single-frame data. Specifically, we exclude the Mamba module and directly regress the extrinsic parameters from the calibration flow map. This approach allows the network to focus on accurately predicting the calibration flow map. We set the initial learning rate to 0.00003 and train for 70 epochs, after which the learning rate is reduced until convergence. All available paired data is utilized in this stage.

In the second stage, we train the Mamba module based on the first stage, enabling the network to fuse multiple calibration flow maps generated over time. The initial learning rate is set to 0.0001 and decays during training. We begin by freezing the network components preceding the Mamba module and then jointly update all network parameters. To enhance training data quality, we use data where the horizontal distance between vehicles and infrastructure is less than 50 meters.

\subsection{Evaluation}

During evaluation, we performed distance filtering before inputting the data into the network. The calibration system is activated only when the vehicle is within certain distance. We will explain the benefits of this operation in Sec.~\ref{sec:4.5}.

\begin{figure}[h]
    \centering
    
    \begin{subfigure}{0.31\linewidth}
        \centering
        \includegraphics[width=\linewidth]{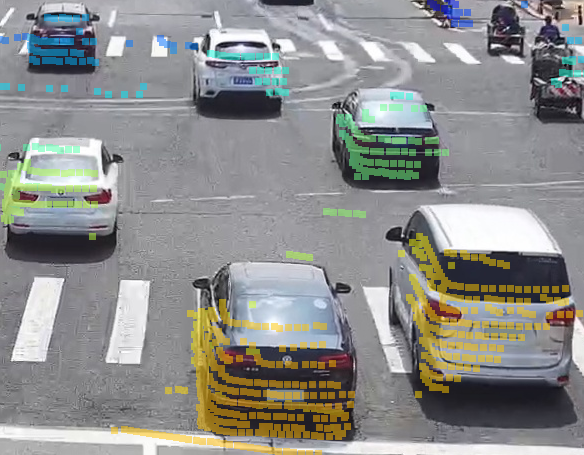}
        
    \end{subfigure}
    \begin{subfigure}{0.31\linewidth}
        \centering
        \includegraphics[width=\linewidth]{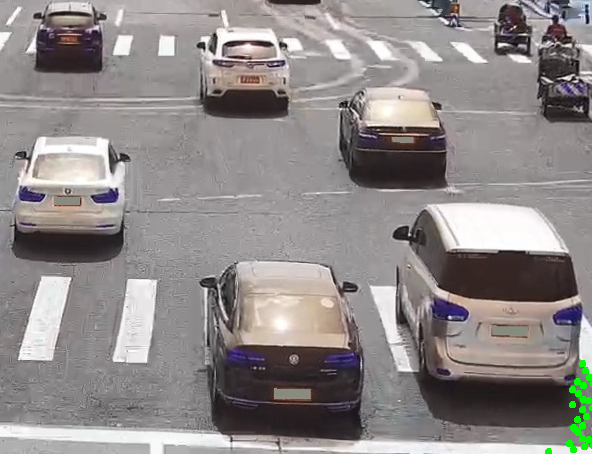}
        
    \end{subfigure}
    \begin{subfigure}{0.31\linewidth}
        \centering
        \includegraphics[width=\linewidth]{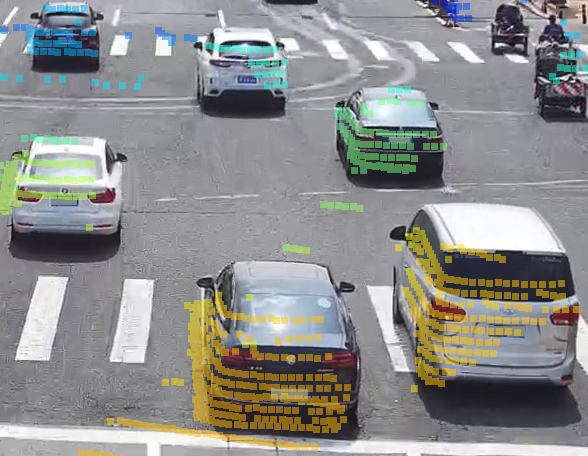}
        
    \end{subfigure}
    
    \caption{Ground Truth (Left). Calib-Anything~\cite{luo2023calib} (Middle) fails to calibrate within (-20°, +20°), while ours (Right) succeeds.}
    
    \label{fig:calibany}
   
\end{figure}

\textbf{On V2X-Seq Dataset~\cite{yu2023v2x}:} Tab.~\ref{tab:my_table} highlights the effectiveness of our method in roadside camera calibration tasks. For a mis-calibration range of $(-20^\circ, +20^\circ)$, our approach achieves a mean calibration error of $0.6313^\circ$, with mean Euler angle errors of $[0.1867^\circ, 0.2409^\circ, 0.4780^\circ]$. We compare our method to a high-performing single-vehicle LiDAR-camera calibration technique~\cite{lv2021lccnet}, demonstrating that our approach offers superior accuracy in V2X scenarios over directly applying the single-vehicle method. Furthermore, we assess its robustness. Our method significantly reduces the likelihood of extreme calibration failures, lowering the standard deviation of the angle-error from $2.5336^\circ$ to $0.3211^\circ$—a reduction by an order of magnitude. The deviations within this range are manageable by existing perception methods~\cite{yang2023bevheight}. When the initial error is reduced to $(-10^\circ, +10^\circ)$, our method yields even better results while maintaining greater robustness and a lower risk of severe errors. Additionally, we present a qualitative comparison with the latest train-free LiDAR-camera calibration method~\cite{luo2023calib} in Fig.~\ref{fig:calibany}. In V2X scenarios with substantial camera deflection, the competing method fails, whereas our approach consistently succeeds. We also compared with a traditional LiDAR-camera calibration method~\cite{koide2023general} (see Appendix 4.3 for details).

\textbf{On TUMTraf-V2X Dataset~\cite{zimmer2024tumtraf}:} Tab.~\ref{TAB:TUMTraf} presents comparative results on the TUMTraf-V2X dataset, where our method remains competitive. For a mis-calibration range of $(-5^\circ, +5^\circ)$, our approach achieves a mean calibration error of $0.267 ^\circ$. Remarkably, our calibration strategy proves effective even in nighttime scenarios (see Fig.~\ref{fig:comparison}).

To ensure a fair comparison, we retrained LCCNet~\cite{lv2021lccnet} on the same dataset and modified its output to predict only rotational errors, aligning it with our network's configuration. Fig.~\ref{fig:comparison} shows that after passing through our calibration network, effective calibration can still be achieved despite significant initial extrinsic parameter deviations. Based on the calibrated position of the roadside camera, the vehicle point clouds can be well-matched with the images from the perspective of the roadside camera.

\begin{table}[htbp]
    
    \centering
    \begin{tabular}{cccc}
    \hline
Deviation & Method & Mean ($^\circ$) & Std ($^\circ$) \\
 \hline
    $(-5^\circ, +5^\circ)$   & LCCNet~\cite{lv2021lccnet}   & 0.389 & 0.608  \\ 
    $(-5^\circ, +5^\circ)$   & Ours   & \bf{0.267} & \bf{0.280}  \\\hline
    \end{tabular}
    
    \caption{Comparison results on TUMTraf-V2X dataset~\cite{zimmer2024tumtraf}.}
    \label{TAB:TUMTraf}
    
\end{table}

\subsection{Ablations and Analysis}
We conducted a series of ablation experiments to demonstrate the importance of each component.

\textbf{Mamba}: First, we removed the Mamba component used for spatiotemporal fusion while retaining feature flow estimation for direct regression. Results
are shown in Tab.~\ref{tab:Ablation}.  It can be observed that the Mamba module significantly improves calibration stability, greatly reducing the standard deviation of the estimated values, which indicates the necessity of this module. Additionally, we attempted to replace it with a Transformer-based Timesformer~\cite{bertasius2021space} module. It can be seen that the Mamba-based method achieves relatively better results with a smaller number of parameters.

\textbf{Iteration}: Our method refines the calibration flow progressively through iteration. During inference, we iterate 10 times to generate 10 calibration flow maps, which together provide spatial dimensional information. We verified that using only a single iteration to produce the calibration flow during inference significantly weakens the network’s capability, indicating that generating calibration flow maps through multiple iterations is essential.

\textbf{Multi-scale 4D Correlation Volume}: We replace the 4D correlation volume with the 3D feature matching layer in LCCNet~\cite{lv2021lccnet}, and find it costs much more parameters and perform worse than 4D correlation volume with iteration (marked by w/o Temporal). 

\subsection{Other experiments about V2X Calibration}
\label{sec:4.5}

\begin{figure}[htbp]
    \centering
    
    \begin{subfigure}{0.31\linewidth}
        \centering
        \includegraphics[width=\linewidth]{re_image/image1.png}
        
    \end{subfigure}
    \begin{subfigure}{0.31\linewidth}
        \centering
        \includegraphics[width=\linewidth]{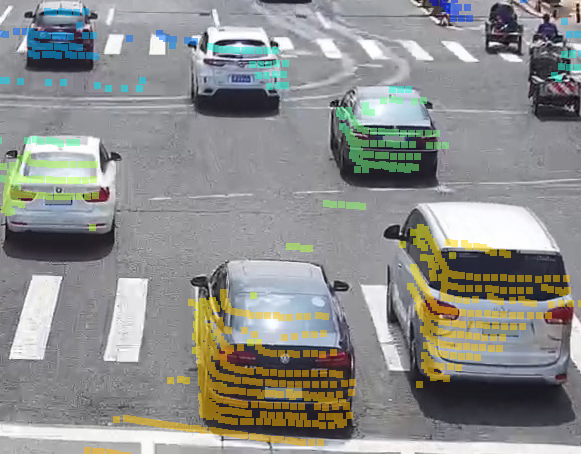}
        
    \end{subfigure}
    \begin{subfigure}{0.31\linewidth}
        \centering
        \includegraphics[width=\linewidth]{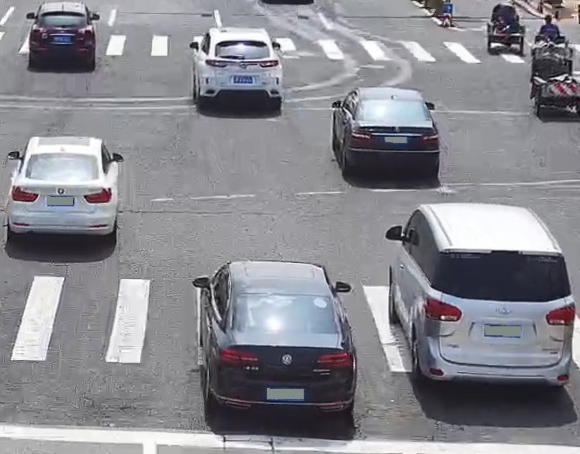}
        
    \end{subfigure}
    
    \caption{The impact of extrinsic noise. Ground Truth (Left). Translated within 10 cm (Middle). Rotated within 20° (Right).}
    
    \label{fig:trans}
\end{figure}

\textbf{Why Ignore Translation Noise:} Fig.~\ref{fig:trans} shows the point cloud projection deviations caused by translational and rotational errors in a V2X scenario. It is evident that minor translational have a negligible impact on the perception system and can be disregarded. In contrast, rotational errors often lead to significant matching discrepancies in vehicle-road data. The experiment in the appendix demonstrate that such slight deviations do not significantly affect rotational calibration. We attribute this to the fact that, in V2X scenarios, the absolute distance between the vehicle and the road typically far exceeds the potential translational offset of roadside cameras due to environmental factors. Therefore, it is reasonable to simplify the calibration problem of roadside cameras by focusing on rotation.

\textbf{Necessity of the Mamba Module:} To further substantiate the importance of the Mamba module, we expanded our ablation study by comparing pre-fusion and post-fusion approaches. As shown in Tab.~\ref{TAB:Fusion}, on the V2X-Seq dataset~\cite{yu2023v2x}, our method markedly outperforms two alternatives: directly stacking multi-frame point clouds or applying post-processing to the frames using a single-frame technique.

\begin{figure}[ht]
    \centering
    \includegraphics[width=0.9\linewidth]{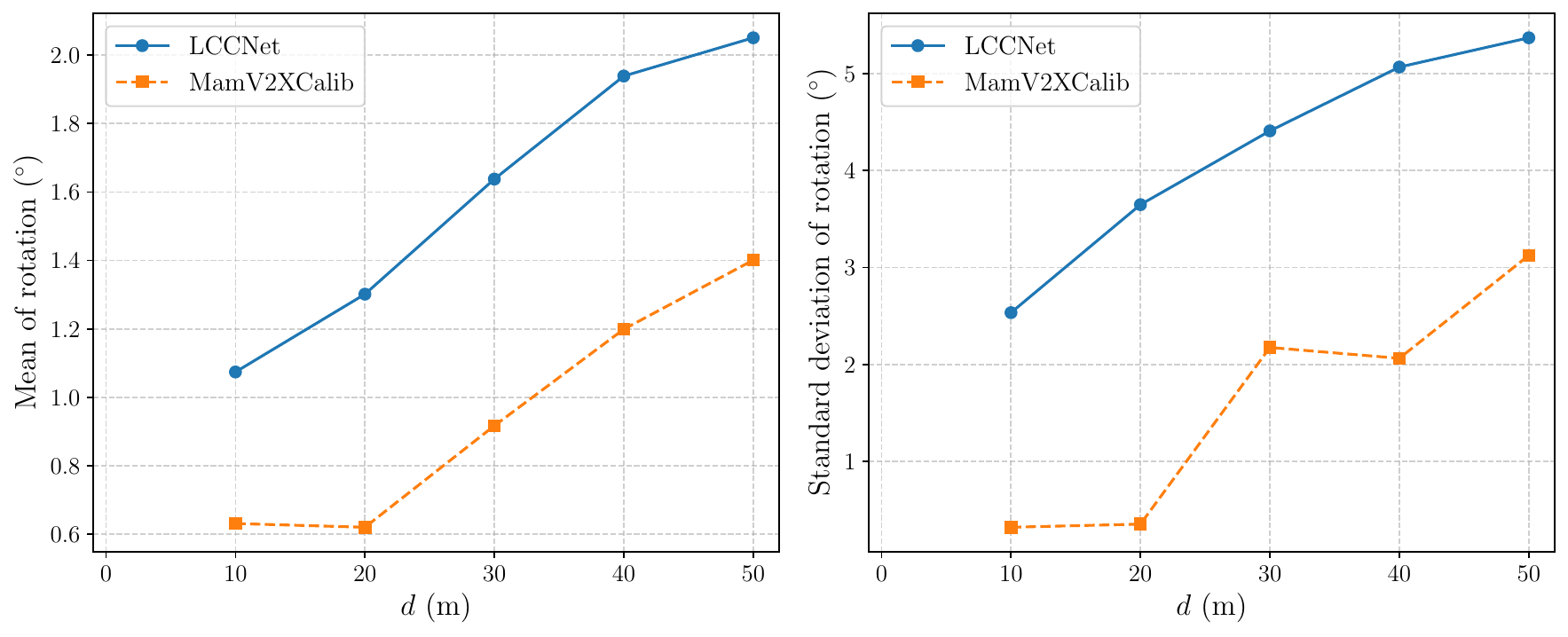}
    \caption{Calibration result on V2X-Seq  dataset~\cite{yu2023v2x} after filtering vehicle point cloud data with a horizontal distance between the vehicle and roadside camera of distance $< d$.}
    \label{fig:fliter}
\end{figure}

\textbf{Distance Filtering:} By utilizing vehicle mobility, we can easily choose data with shorter distances between vehicles and infrastructure for calibration. This distance can be calculated before feeding it into the neural network, allowing us to initially evaluate data quality based on distance to boost calibration accuracy. This is an advantage that earlier methods lack. In Fig.~\ref{fig:fliter}, we show the mean error and standard deviation of calibration using vehicle point clouds across various distance ranges. Experiments demonstrate that filtering data by distance improves data quality and strengthens the method's robustness.

\textbf{Migration to single car calibration:} Although our method was originally designed for V2X, it can naturally be extended to single car where translation errors must be considered. The results are shown in the attachment.

\begin{table}[htbp]
    
    \centering
    \begin{tabular}{ccc}
    \hline
Fusion Methods  & Mean ($^\circ$) & Std ($^\circ$) \\
 \hline
    stacking points    & 0.952 & 2.184  \\ 
    average    & 0.974 & 2.001  \\
    least square fitting    & 0.820   & 1.254  \\
    ours   &  \bf{0.631}   &  \bf{0.321} \\\hline
    \end{tabular}
    
    \caption{Comparison with other fusion methods on the V2X-Seq dataset~\cite{yu2023v2x} (initial deviation $(-20^\circ, +20^\circ)$).}
    \label{TAB:Fusion}
    
\end{table}

\subsection{Limitation}
Our method relies on vehicle-side LiDAR perception data, which places certain demands on the quality of vehicle-side sensors. Additionally, although the V2X-based calibration strategy offers benefits such as target-free operation, high flexibility, and large-scale deployment, it also introduces vehicle-to-infrastructure information interaction, which imposes higher requirements on time synchronization. Besides, similar to other deep learning-based calibration methods, fine-tuning is necessary to achieve good results when transferring to entirely different datasets.

\section{Conclusion}
This paper proposes a target-less infrastructure camera calibration strategy based on vehicle-road collaboration and introduces a novel method designed specifically for V2X scenarios. Compare to existing approaches, our strategy makes full use of environmental information and is suited for busy road scenarios. We present an iterative method that establishes pixel-level correspondences between depth maps and images. By incorporating the Mamba model’s strengths in temporal modeling, we overcome instability challenges in single-vehicle calibration applied to V2X contexts with fewer parameters. Our method proves effective and robust on two datasets. We hope that it can inspire future related study.
\section*{Acknowledgements}
This work is supported by National Science and Technology Major Project (2022ZD0115502), and Wuxi Research Institute of Applied Technologies, Tsinghua University under Grant 20242001120.\\
{
    \small
    \bibliographystyle{ieeenat_fullname}
    \bibliography{main}
}

\end{document}